\pdfoutput=1

\documentclass[11pt]{article}

\usepackage{naacl2021}

\usepackage{times}
\usepackage{latexsym}
\usepackage{subcaption}
\usepackage{soul}
\usepackage[T1]{fontenc}

\usepackage[utf8]{inputenc}

\usepackage{microtype}

\usepackage{amssymb}
\usepackage{derivative}
\usepackage{url}
\usepackage{mathtools,amsthm}

\usepackage{bm}
\usepackage{threeparttable}
\usepackage{multirow}
\usepackage[ruled,vlined]{algorithm2e}
\usepackage{caption}
\usepackage{subcaption}
\usepackage{pbox}
\usepackage{booktabs}
\newcommand{\citeA}[1]{\citeauthor{#1} (\citeyear{#1})}
\newcommand{\ptacek}{Pt{\'a}{\v{c}}ek}

\newcommand\blfootnote[1]{%
  \begingroup
  \renewcommand\thefootnote{}\footnote{#1}%
  \addtocounter{footnote}{-1}%
  \endgroup
}

\title{Latent-Optimized Adversarial Neural Transfer for Sarcasm Detection}

\author{Xu Guo, Boyang Li*, Han Yu \and Chunyan Miao*\\
  School of Computer Science and Engineering,\\
  Nanyang Technological University, Singapore\\
  \texttt{\{xu008, han.yu, boyang.li, ascymiao\}@ntu.edu.sg} \\
}

\begin{document}
\maketitle

\begin{abstract}
The existence of multiple datasets for sarcasm detection prompts us to apply transfer learning to exploit their commonality. The adversarial neural transfer (ANT) framework utilizes multiple loss terms that encourage the source-domain and the target-domain feature distributions to be similar while optimizing for domain-specific performance. However, these objectives may be in conflict, which can lead to optimization difficulties and sometimes diminished transfer. We propose a generalized latent optimization strategy that allows different losses to accommodate each other and improves training dynamics. The proposed method outperforms transfer learning and meta-learning baselines. In particular, we achieve 10.02\% absolute performance gain over the previous state of the art on the iSarcasm dataset. 
\end{abstract}

\section{Introduction\blfootnote{* Corresponding authors}}

Sarcastic language is commonly found in social media posts \cite{ibanez2011:sarcasm-in-twitter,maynard2014cares}, forum discussions \cite{khodak2018large}, product reviews \cite{Davidov2010:Twitter-Amazon,filatova-2012-irony} and everyday conversations \cite{Gibbs2000:Irony-in-Talk}. Detecting sarcasm is an integral part of creative language understanding \cite{veale2019systematizing} and online opinion mining \cite{kannangara2018mining}. Due to highly contextualized expressions, detecting sarcasm is a challenging task, even for humans \cite{FoxTree2020}. 


A challenge specific to sarcasm detection is the difficulty in acquiring ground-truth annotations. Human-annotated datasets \cite{filatova-2012-irony,Riloff2013,van2018semeval,oprea-magdy:2020:isarcasm} usually contain only a few thousand texts, resulting in many small datasets. In comparison, automatic data collection using distant supervision signals like hashtags \cite{ptavcek2014sarcasm,bamman2015contextualized,joshi2015harnessing} yielded substantially larger datasets. Nevertheless, the automatic approach also led to label noise. For example, \citeA{oprea-magdy:2020:isarcasm} found nearly half of the tweets with sarcasm hashtags in one dataset are not sarcastic. 

The existence of diverse datasets and data collection methods prompts us to exploit their commonality through transfer learning. Specifically, we transfer knowledge learned from large and noisy datasets to improve sarcasm detection on small human-annotated datasets that serve as effective performance benchmarks.  



Adversarial neural transfer (ANT) \cite{Ganin2015,liu-etal-2017-adversarial,kim2017adversarial,kamath2019reversing} employs an adversarial setup where the network learns to make the shared feature distributions of the source domain and the target domain as similar as possible, while simultaneously optimizing for domain-specific performance. However, as the domain-specific losses promote the use of domain-specific features, these training objectives may compete with each other implicitly. This leads to optimization difficulties and potentially degenerate cases where the domain-specific classifiers ignore the shared features and no meaningful transfer occurs between domains. 

To cope with this issue, we propose Latent-Optimized Adversarial Neural Transfer (LOANT). The latent optimization strategy can be understood with analogies to to one-step look-ahead during gradient descent and Model-Agnostic Meta Learning  \cite{pmlr-v70-finn17a}. By forcing domain-specific losses to accommodate the negative domain discrimination loss, it improves training dynamics \cite{Balduzzi2018:n-Player}.


With LOANT, we achieve 10.02\% absolute improvement over the previous state of the art on the iSarcasm dataset \cite{oprea-magdy:2020:isarcasm} and 3.08\% improvement on SemEval-18 dataset \cite{van2018semeval}. Over four sets of transfer learning experiments, latent optimization on average brings 3.42\% improvement in F-score over traditional adversarial neural transfer and 4.83\% over a similar training strategy from Model-Agnostic Meta Learning (MAML) \cite{pmlr-v70-finn17a}. In contrast, traditional ANT brings an average of only 0.9\% F-score improvement over non-adversarial multi-task learning. 
The results demonstrates that LOANT can effectively perform knowledge transfer for the task of sarcasm detection and suggests that the proposed latent optimization strategy enables the collaboration among the ANT losses during optimization.

Our contributions can be summarized as follows:

\begin{enumerate}
    \item Inspired by the existence of multiple small sarcasm datasets, we propose to use transfer learning to bridge dataset differences. To the best of our knowledge, this is the first study of transfer learning between different sarcasm detection datasets. 
    
    \item We propose LOANT, a novel latent-optimized adversarial neural transfer model for cross-domain sarcasm detection. By conducting stochastic gradient descent (SGD) with one-step look-ahead, LOANT outperforms traditional adversarial neural transfer, multi-task learning, and meta-learning baselines, and establishes a new state-of-the-art F-score of 46.41\%. The code and datasets are available at \url{https://github.com/guoxuxu/LOANT}.
\end{enumerate}

\section{Related Work}
\subsection{Sarcasm Detection}



Acquiring large and reliable datasets has been a persistent challenge for computational detection of sarcasm. Due to the cost of annotation, manually labeled datasets \cite{walker2012corpus,Riloff2013,wallace2014humans,abercrombie2016putting,oraby-etal-2016-creating,van2018semeval,oprea-magdy:2020:isarcasm} typically contain only a few thousand texts. Automatic crawling \cite{ptavcek2014sarcasm,bamman2015contextualized,joshi2015harnessing,khodak-etal-2018-large} using hashtags or markers yields substantially more texts, but the results are understandably more noisy. As a case study, after examining the dataset of \citeA{Riloff2013}, \citeA{oprea-magdy:2020:isarcasm} found that nearly half of tweets with sarcasm hashtags are not sarcastic. In this paper, we evaluate performance on the manually labeled datasets, which are relatively clean and can serve as good benchmarks, and transfer the knowledge learned from automatically collected datasets.

Traditional sarcasm detection includes methods based on rules \cite{tepperman2006yeah} and lexical \cite{kreuz-caucci-2007-lexical} and pragmatic patterns \cite{gonzalez2011identifying}. Context-aware methods \cite{rajadesingan2015sarcasm,bamman2015contextualized} make use of contexts, such as the author, the audience, and the environment, to enrich feature representations. 

Deep learning techniques for sarcasm detection employ convolutional networks \cite{ghosh2016fracking}, recurrent neural networks  \cite{zhang2016tweet,felbo-etal-2017-using,wu2018thu_ngn}, attention  \cite{tay-etal-2018-reasoning}, and pooling \cite{xiong2019sarcasm} operations. \citeA{amir2016modelling} incorporate historic information for each Twitter user. \citeA{cai-etal-2019-multi} consider the images that accompany tweets and \citeA{mishra-2017-gaze} utilize readers' gaze patterns. To the best of our knowledge, no prior work has explored transfer learning between different sarcasm datasets.

\subsection{Adversarial Transfer Learning}

As a transfer learning technique, multi-task learning (MTL) allows related tasks or similar domains to inform each other and has been a powerful technique for NLP \cite{Collobert2011,yang2017transfer,aharoni-etal-2019-massively,guo2019autosem,Raffel2020:T5}. However, MTL does not always lead to performance improvements \cite{alonso2017multitask,bingel-sogaard-2017-identifying,changpinyo2018multitask,clark2019bam}.

Theoretical analysis \cite{ben2010theory} indicates that a key factor for the success of transfer is to reduce the divergence between the feature spaces of the domains. \citeA{Ganin2015} propose to minimize domain differences via a GAN-like setup, where a domain discriminator network learns to distinguish between features from two domains and a feature extraction network learns to produce indistinguishable features, which are conducive to transfer learning.

Similar adversarial setups \cite{liu-etal-2017-adversarial,kim2017adversarial} have been adopted for many NLP tasks, such as sentiment analysis \cite{chen2018adversarial,liu2018learning}, satirical news detection \cite{mchardy-etal-2019-adversarial}, detection of duplicate questions \cite{kamath2019reversing}, named entity recognition \cite{zhou2019dual_c}, and QA \cite{YuJianfei2018}. 

However, as shown in our experiments, adding the domain discriminator to MTL does not always result in improved performance. We attribute this to the implicit competition between the negative domain discrimination loss and the domain-specific losses, which causes difficulties in optimization. In this paper, we improve the training dynamics of adversarial transfer learning using latent optimization on BERT features.


\subsection{Meta-Learning and Latent Optimization}
The idea of coordinating gradient updates of different and competing losses using gradient descent with look-ahead has been explored in Latent-optimized Generative Adversarial Network (LOGAN) \cite{pmlr-v97-wu19d,wu2019logan}, Symplectic Gradient Adjustment \cite{Balduzzi2018:n-Player,gemp2019global}, Unrolled GAN \cite{metz2016unrolled}, Model-Agnostic Meta Learning \cite{pmlr-v70-finn17a} and extragradient \cite{azizian2020:extragradient}. The difference between LOGAN and other techniques is that the LOGAN computes the derivative of the randomly sampled latent input, whereas other methods compute the second-order derivative in the model parameter space. 

In this paper, we generalize latent optimization from GANs to multi-task learning, where the adversarial loss is complemented by domain-specific task losses. In addition, we apply latent optimization on the output of the BERT module, which differs from the optimization of the random latent variable in LOGAN. As large pretrained masked language models (PMLMs) gain prominence in NLP, latent optimization avoids gradient computation on the parameters of enormous PMLMs, providing reduction in running time and memory usage.

\section{The LOANT Method}
In supervised transfer learning, we assume labeled data for both the source domain and the target domain are available. The source domain dataset $D_s$ comprises of data points in the format of $(x_s, y_s)$ and the target domain dataset $D_t$ comprises of data points in the format of $(x_t, y_t)$. The labels $y_s$ and $y_t$ are one-hot vectors. The task of supervised cross-domain sarcasm detection can be formulated as learning a target-domain function $f_t(x_t)$ that predict correct labels for unseen $x_t$. 


\begin{figure}[t]
    \centering
    \includegraphics[width=0.8\columnwidth]{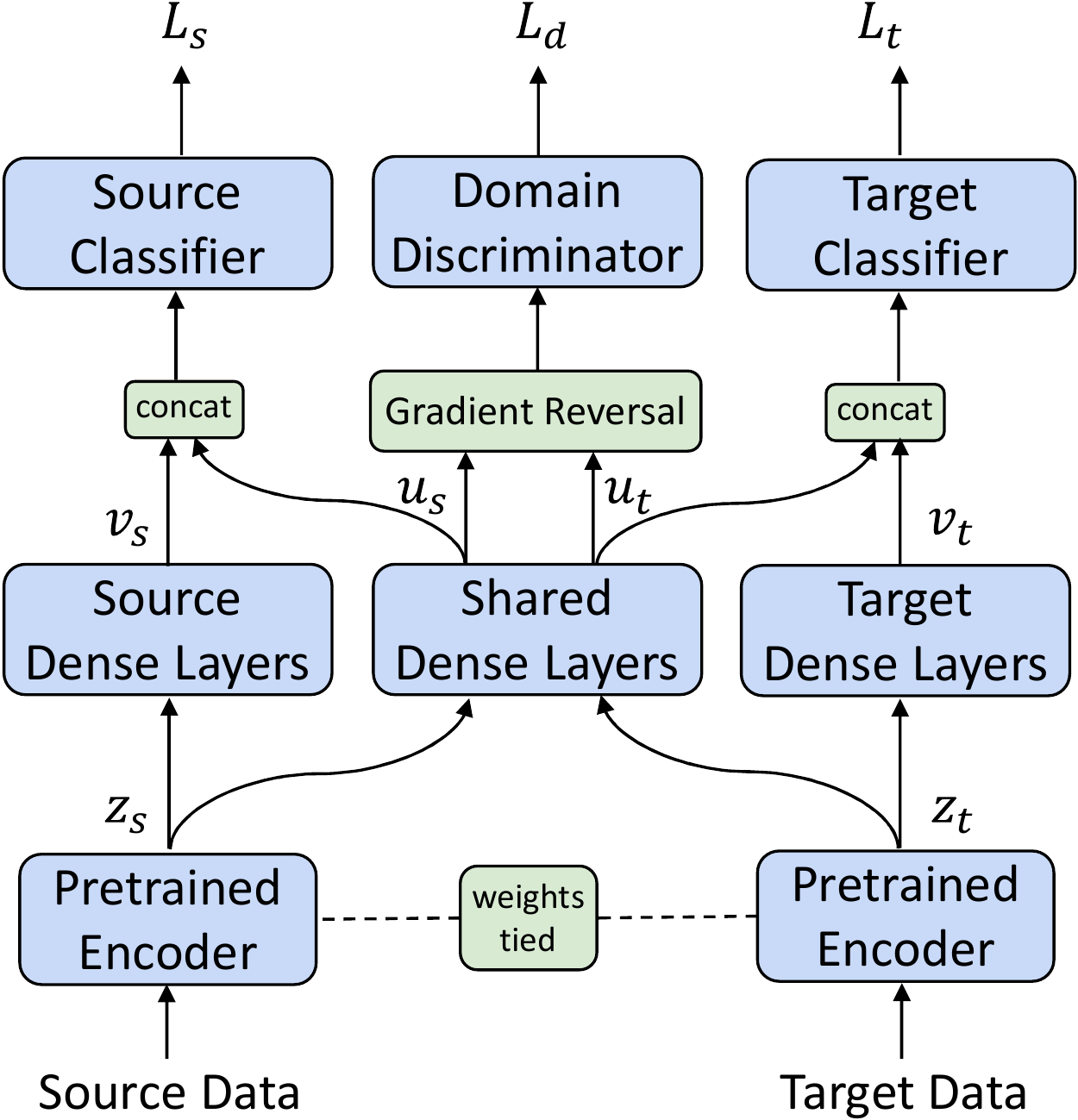}
    \caption{Network architecture of the Adversarial Neural Transfer model.}
    \label{fig:LOANT_architecture}
\end{figure}

\subsection{Model Architecture}

Fig. \ref{fig:LOANT_architecture} shows the model architecture for adversarial neural transfer (ANT) \cite{liu-etal-2017-adversarial,kamath2019reversing,kim2017adversarial}. We use a large pretrained neural network, BERT \cite{devlin-etal-2019-bert}, as the sentence encoder, though the architecture is not tied to BERT and can use other pretrained encoders. We denote the parameters of the BERT encoder as $w_b$, and its output for data in the source domain and the target domain as $z_s \in \mathbb{R}^{D}$ and $z_t \in \mathbb{R}^{D}$ respectively. We denote this encoder operation as
\begin{equation}
    z_s = E(x_s, w_b), \, z_t = E(x_t, w_b)
\end{equation}
On top of these outputs, we apply domain-specific dense layers to create domain-specific features $v_s, v_t$ and shared dense layers to create shared features $u_s, u_t$. We use $w_s$, $w_t$, and $w_{sh}$ to denote the parameters for the source dense layers, the target dense layers, and the shared dense layers. 

The concatenation of features $[v_s, u_s]$ is fed to the source-domain classifier, parameterized by $\theta_s$; $[v_t, u_t]$ is fed to the target-domain classifier, parameterized by $\theta_t$. The two classifiers categorize the tweets into sarcastic and non-sarcastic and are trained using cross-entropy. For reasons that will become apparent later, we make explicit the reliance on $z_s$ and $z_t$:
\begin{equation}
\begin{split}
    \mathcal{L}_s(z_s) & = - \sum_i y_{s,i} \log  p(\hat{y}_{s,i}|z_s),
    \\
    \mathcal{L}_t(z_t) & = - \sum_i y_{t,i} \log p(\hat{y}_{t,i}|z_t),
\end{split}
\end{equation}
where $\hat{y}_s$ and $\hat{y}_t$ are the predicted labels and $i$ is the index of the vector components. 

Simultaneously, the domain discriminator learns to distinguish the features $u_s$ and $u_t$ as coming from different domains. The domain discriminator is parameterized by $\theta_d$. It is trained to minimize the domain classification loss,
\begin{equation}
    \mathcal{L}_d(z_t, z_s) = - \log p(0|u_s)  - \log p(1|u_t).
\end{equation}
Through the use of the gradient reversal layer, the shared dense layers and the feature encoder maximizes the domain classification loss, so that the shared features $u_s$ and $u_t$ become indistinguishable and conducive to transfer learning. In summary, the network weights $w_b, w_s, w_t, w_{sh}, \theta_s, \theta_t$ are trained to minimize the following joint loss,
\begin{equation}
\label{eq:shared-loss}
    \mathcal{L}^{\text{ANT}} = \mathcal{L}_s(z_s) + \mathcal{L}_t(z_t) - \mathcal{L}_d(z_t, z_s),
\end{equation}
whereas $\theta_d$ is trained to minimize $\mathcal{L}_d(z_t, z_s)$. 

It is worth noting that the effects of three loss terms in Eq. \ref{eq:shared-loss} on the shared parameters $w_{sh}$ and $w_b$ may be competing with each other. This is because optimizing sarcasm detection in one domain will encourage the network to extract domain-specific features, whereas the domain discrimination loss constrains the network to avoid such features. It is possible for the competition to result in degenerate scenarios. For example, the shared features $u_s$ and $u_t$ may become indistinguishable but also do not correlate with the labels $y_s$ and $y_t$. The domain classifiers may ignore the shared features $u_s$ and $u_t$ and hence no transfer happens. To cope with this issue, we introduce a latent optimization strategy that forces domain-specific losses to accommodate the domain discrimination loss.

\begin{figure}[t]
    \centering
    \includegraphics[width=0.8\columnwidth]{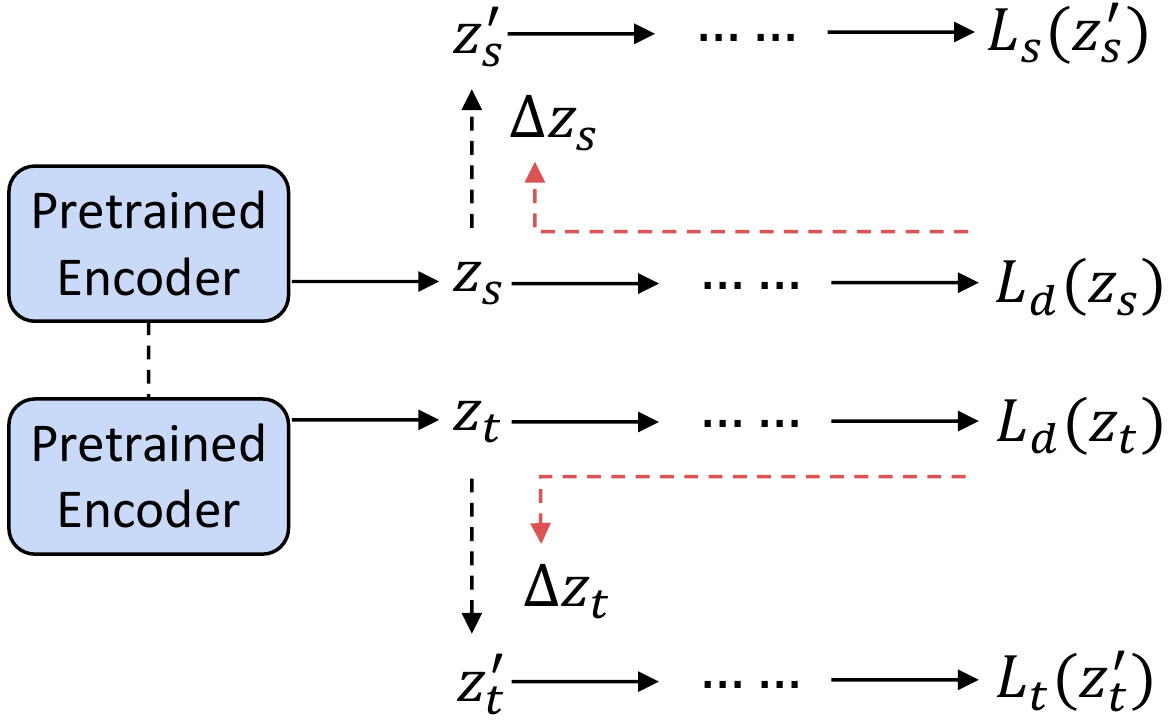}
    \caption{Schematic of the latent optimization strategy. The solid black arrows indicate the forward pass and the dotted red arrows indicate the backward pass.}
    \label{fig:lo}
\end{figure}

\subsection{Latent Representation Optimization}


We now introduce the latent representation optimization strategy. First, we perform one step of stochastic gradient descent on $-\mathcal{L}_d$ on the encoded features $z_s$ and $z_t$ with learning rate $\gamma$,
\begin{align}
\label{eq:loant-look-ahead-1}
z_{s}^\prime & = z_{s} + \gamma \pdv{\mathcal{L}_d(z_s, z_t)}{z_s}, \\
\label{eq:loant-look-ahead-2}
z_{t}^\prime & = z_{t} + \gamma \pdv{\mathcal{L}_d(z_s, z_t)}{z_t}.
\end{align}
We emphasize that this is a \emph{descent} step because we are minimizing $-\mathcal{L}_d$.

After that, we use the updated $z_{s}^\prime$ and $z_{t}^\prime$ in the computation of the losses and the new joint objective hence becomes
\begin{equation}
\label{eq:new_joint_loss}
\mathcal{L}^{\text{LO}} = \mathcal{L}_s(z_s^\prime) + \mathcal{L}_t(z_t^\prime)  - \mathcal{L}_d(z_s, z_t),
\end{equation}
which is optimized using regular stochastic gradient descent (SGD) on $w_b, w_s, w_t, w_{sh}, \theta_s$, and $\theta_t$. 



Here we show the general case of gradient computation. Consider any weight vector $w$ in the neural network. Equations \ref{eq:loant-look-ahead-1} and \ref{eq:loant-look-ahead-2} introduce two intermediate variables $z_s^\prime$ and $z_t^\prime$, which are a function of the model parameter $w$. Therefore, we perform SGD using the following total derivative
\begin{equation} \label{expansion-1}
\begin{split}
\odv{\mathcal{L}^{\text{LO}}}{w} =  \pdv{\mathcal{L}^{\text{LO}}}{w} + \pdv{\mathcal{L}_{s}(z_{s}^\prime)}{z_{s}^\prime}\pdv{z_{s}^\prime}{w} + \pdv{\mathcal{L}_{t}(z_{t}^\prime)}{z_{t}^\prime}\pdv{z_{t}^\prime}{w}.
\end{split}
\end{equation}
where
\begin{equation}\label{expansion-2}
\begin{split}
\pdv{z_{s}^\prime}{w}=\pdv{z_{s}}{w} + \gamma \pdv{\mathcal{L}_{d}(z_{s})}{z_{s},w}
\\
\pdv{z_{t}^\prime}{w}=\pdv{z_{t}}{w} + \gamma \pdv{\mathcal{L}_{d}(z_{t})}{z_{t},w}
\end{split}
\end{equation}
For every network parameter other than the encoder weight $w_b$, $\pdv{z}/{w}$ is zero. The second-order derivative $\pdv{\mathcal{L}_d(z)}/{z,w}$ is difficult to compute due to the high dimensionality of $w$. Since $\gamma$ is usually very small, we adopt a first-order approximation and directly set the second-order derivative to zero. 
Letting $\phi_s = [w_s, \theta_s]$ and $\phi_t = [w_t, \theta_t]$, we now show the total derivatives for all network parameters:
\begin{align}
\begin{split}
\odv{\mathcal{L}^{\text{LO}}}{w_b} & = \pdv{\mathcal{L}_s(z_s^\prime)}{w_b} + \pdv{\mathcal{L}_t(z_t^\prime)}{w_b} \\
& + \pdv{\mathcal{L}_s(z_s^\prime)}{z_s^\prime} \pdv{z_s}{w_b} + \pdv{\mathcal{L}_t(z_t^\prime)}{z_t^\prime} \pdv{z_t}{w_b}
\end{split}\\
\begin{split}
\odv{\mathcal{L}^{\text{LO}}}{w_{sh}} & = \pdv{\mathcal{L}_s(z_s^\prime)}{w_{sh}} + \pdv{\mathcal{L}_t(z_t^\prime)}{w_{sh}}
\end{split}\\ 
\begin{split}
\odv{\mathcal{L}^{\text{LO}}}{ \phi_{s}} & = \pdv{\mathcal{L}_s(z_s^\prime)}{\phi_{s}} \\
\odv{\mathcal{L}^{\text{LO}}}{\phi_{t}} & = \pdv{\mathcal{L}_t(z_t^\prime)}{\phi_{t}} \\
\odv{\mathcal{L}^{\text{LO}}}{\theta_d} & = \pdv{\mathcal{L}_d(z_s, z_t)}{\theta_d}
\end{split}
\end{align}
More details can be found in Appendix \ref{appendix:First_order_Approximation}. Fig. \ref{fig:lo} illustrates the latent optimization process. Algorithm \ref{algo:training_LOANT} shows the LOANT algorithm.

\begin{algorithm}[t]
\SetAlgoVlined
\DontPrintSemicolon
\KwIn{source data $(x_s, y_s)$, target data $(x_t, y_t)$, learning rate $\gamma$\;}
 Initialize model parameters $w$\;
 \Repeat{the maximum training epoch}{
  Sample $N$ batches of data pairs\;
  \For{i = 1 \KwTo $N$}{
      Compute forward cross entropy loss $\mathcal{L}_s(z_s)$, $\mathcal{L}_t(z_t)$, $\mathcal{L}_d(z_s, z_t)$;\\
      Compute $\bigtriangleup z_s=\pdv{\mathcal{L}_d(z_s, z_t)}{z_s}$ and $\bigtriangleup z_t=\pdv{\mathcal{L}_d(z_s, z_t)}{z_t}$;\\
      Update the latent representations $z_s^\prime=z_s +\gamma\bigtriangleup z_s$ and $z_t^\prime=z_t +\gamma\bigtriangleup z_t$;\\
      Compute the new joint loss
      $\mathcal{L}^{\text{LO}} = \mathcal{L}_s(z_s^\prime) + \mathcal{L}_t(z_t^\prime) - \mathcal{L}_d(z_s, z_t)$;\\
      Update $w$ using gradient descent.
  }
 }
 \caption{Training of LOANT}
 \label{algo:training_LOANT}
\end{algorithm}

\subsection{Understanding LOANT}
\label{sec:rationale-for-loant}
To better understand the LOANT algorithm, we relate LOANT to the extragradient technique and Model-Agnostic Meta Learning \cite{pmlr-v70-finn17a}.

\begin{figure*}
    \centering
    \begin{subfigure}{.32\textwidth}
  \centering
  \includegraphics[width=\textwidth]{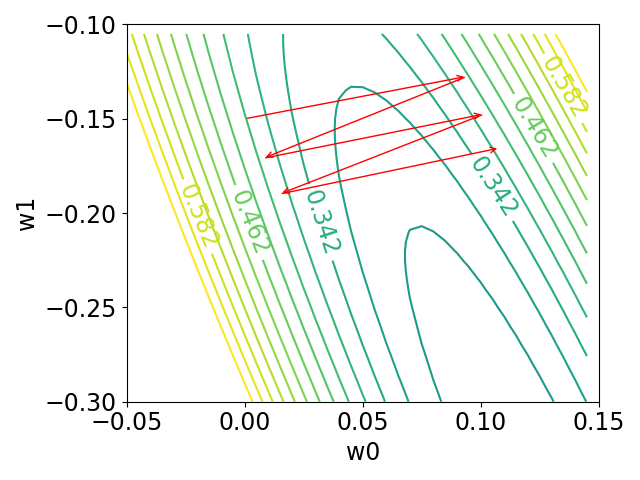}
  \caption{Vanilla gradient descent, which exhibits a zigzag trajectory. $\eta=0.025$.\\}
  \label{fig:gradient_descent}
\end{subfigure}%
\hfill 
\begin{subfigure}{.32\textwidth}
  \centering
  \includegraphics[width=\textwidth]{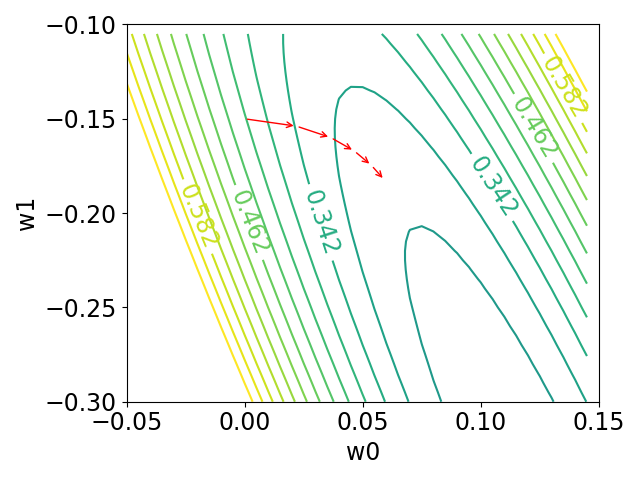}
  \caption{First-order extragradient, which sets the Hessian term to zero. $\eta=0.025$. $\gamma=0.01$.}
  \label{fig:gd_first_order_LookAhead}
\end{subfigure}
\hfill 
\begin{subfigure}{.32\textwidth}
  \centering
  \includegraphics[width=\textwidth]{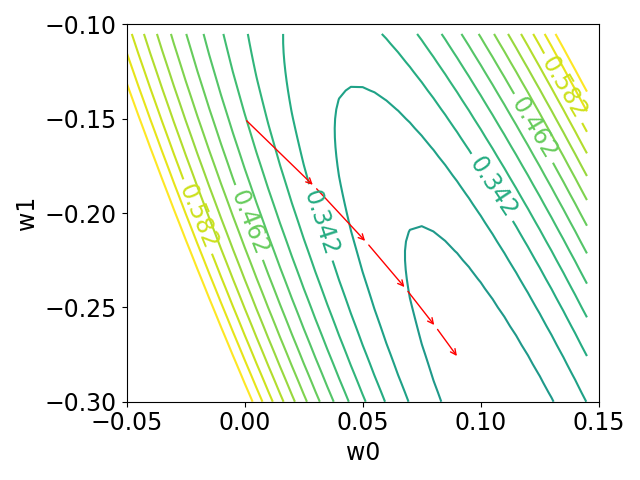}
  \caption{Full-Hessian extragradient, which finds a direct path to the local minimum, enabling a large learning rate $\eta=0.1$. }
  \label{fig:gd_second_order_LookAhead}
\end{subfigure}
\caption{Minimization of a 2D function $f(w) = w^{\top}Aw + b^{\top}w + c$. $A$ is positive definite and has a condition number of 40. The initial point is $(0, -0.15)$. The red arrows show the trajectory of $w$. The look-ahead capability of extragradient finds a much more direct path to the local minimum than vanilla gradient descent.}
\label{fig:toy_example}
\end{figure*}

The vanilla gradient descent (GD) algorithm follows the direction along which the function value decreases the fastest. However, when facing an ill-conditioned problem like the one in Fig. \ref{fig:toy_example}, GD is known to exhibit slow convergence because the local gradients are close to being orthogonal to the direction of the local optimum.

For comparison with LOANT, we consider the extragradient (EG) method \cite{Korpelevich1976:extragradient,azizian2020:extragradient} that uses the following update rule when optimizing the function $f(w)$ with respect to $w$,
\begin{equation}
\label{eq:extragradient}
    w \leftarrow w - \eta \odv{f(w - \gamma \pdv{f(w)}{w})}{w}.
\end{equation}
Similar to LOANT, we can adopt a first-order approximation to EG if we set the Hessian term to zero in the total derivative. 
Instead of optimizing the immediate function value $f(w)$, this method optimizes $f(w - \gamma\pdv{f}{w})$, which is the function value after one more GD step. 
This can be understood as looking one step ahead along the optimization trajectory. In the contour diagrams of Fig. \ref{fig:toy_example}, we show the optimization of a 2-dimensional quadratic function. This simple example showcases how the ability to look one step ahead can improve optimization in pathological loss landscapes. We motivate the nested optimization of LOANT by drawing an analogy between EG and LOANT.


It is worth noting that LOANT differs from the EG update rule in important ways. Specifically, in EG the inner GD step and the outer GD step are performed on the same function $f(\cdot)$, whereas LOANT performs the inner step on $\mathcal{L}_d$ and the outer step on $\mathcal{L}_s$ or $\mathcal{L}_t$.  

For a similar idea with multiple losses, we turn to MAML \cite{pmlr-v70-finn17a}. In MAML, there are $K$ tasks with losses $\mathcal{L}_1, \ldots, \mathcal{L}_k, \ldots, \mathcal{L}_K$. On every task, we perform a one-step SGD update to the model parameter $w \in \mathbb{R}^L$,
\begin{equation}
\label{eq:maml-look-ahead}
    w_{T_k} = w - \gamma \pdv{\mathcal{L}_k(w)}{w}.
\end{equation}
After going through $K$ tasks, the actual update to $w$ is calculated using the parameters $w_{T_k}$,
\begin{equation}
\label{eq:maml-update}
    w \leftarrow w - \eta \frac{1}{K} \sum_k \odv{\mathcal{L}_k(w_{T_k})}{w}.
\end{equation}
Utilizing the idea of look ahead, in MAML we update $w$ so that subsequent optimization on any single task or combination of tasks would achieve good results. 

Adversarial neural transfer has three tasks, the source-domain and target-domain classifications and the negative discriminator loss. The updates performed by LOANT in Eq. \ref{eq:loant-look-ahead-1} and \ref{eq:loant-look-ahead-2} are similar to MAML's look-ahead update in Eq. \ref{eq:maml-look-ahead}. Specifically, when we update model parameters using the gradient from the total loss $\mathcal{L}^{\text{LO}}$, we prepare for the next descent step on $-\mathcal{L}_d$. Therefore, LOANT can be understood as forcing domain-specific losses to accommodate the domain discrimination loss and mitigating their competition.

LOANT differs from MAML since, in the inner update, LOANT updates the sentence-level features $z_s$ and $z_t$ instead of the model parameters $w$. As $z_s$ and $z_t$ are usually of much smaller dimensions than $w$, this leads to accelerated training and reduced memory footprint. For example, in the BERT-base model \cite{devlin-etal-2019-bert}, $L$ is 110 million and $D$ is 768. Within the regular range of batch size $B$, $BD \ll L$. In the experiments, we verify the benefits of LOANT in terms of accuracy and time and space complexity.




\section{Experiments}


\subsection{Datasets}
We conduct four cross-domain sarcasm detection experiments by transferring from an automatically collected dataset to a manually annotated dataset. The two automatically collected datasets include \ptacek{} \cite{ptavcek2014sarcasm} and Ghosh\footnote{\url{https://github.com/AniSkywalker/SarcasmDetection/tree/master/resource}} \cite{ghosh2016fracking}, which treat tweets having particular hastags such as \texttt{\#sarcastic},  \texttt{\#sarcasm} or  \texttt{\#not} as sarcastic and others as not sarcastic. We crawled the \ptacek{} dataset using the NLTK API\footnote{\url{http://www.nltk.org/howto/twitter.html}} according to the tweet ids published online\footnote{\url{http://liks.fav.zcu.cz/sarcasm/}}. 

The two manually annotated datasets include SemEval-18\footnote{\url{https://github.com/Cyvhee/SemEval2018-Task3/tree/master/datasets}} \cite{van2018semeval} and iSarcasm \cite{oprea-magdy:2020:isarcasm}. SemEval-18 consists of both sarcastic and ironic tweets supervised by third-party annotators and thus is used for \textit{perceived} sarcasm detection. The iSarcasm dataset contains tweets written by participants of an online survey and thus is an example of \textit{intended} sarcasm detection. 

Table \ref{tab:dataset_statistics} summarizes the statistics of the four datasets. The SemEval-18 dataset is balanced while the iSarcasm dataset is imbalanced. The two source datasets are more than ten times the size of the target datasets. For all datasets, we use the predefined test set and use a random 10\% split of the training set as the development set. 


\begin{table}[t]
    \centering
    \small
    \begin{tabular}{lllll}
    \hline
         Dataset & Train & Val & Test & \% Sarcasm \\
         \hline
         \ptacek & 51009 & 5668 & 6298 & 49.50\% \\
         Ghosh & 33373 & 3709 & 4121 & 44.84\% \\
         SemEval-18 & 3398 & 378 & 780 & 49.12\%\\
         iSarcasm & 3116 & 347 & 887 & 17.62\%\\
      \hline
    \end{tabular}
    \caption{Dataset statistics, including number of samples in each split and the proportion of sarcastic texts.}
    \label{tab:dataset_statistics}
\end{table}

We preprocessed all datasets using the lexical normalization tool for tweets from \citeA{baziotis-pelekis-doulkeridis:2017:SemEval2}. We cleaned the four datasets by dropping all the duplicate tweets within and across datasets, and trimmed the texts to a maximum length of 100. To deal with class imbalance, we performed upsampling on the target-domain datasets, so that both the sarcastic and non-sarcastic classes have the same size as source domain datasets.

\subsection{Baselines}
We compare LOANT with several competitive single-task and multi-task baselines. 

\vspace{0.05in}
\noindent\textbf{MIARN} \cite{tay-etal-2018-reasoning}: A state-of-the-art short text sarcasm detection model ranked top-1 on the iSarcam dataset. The model is a co-attention based LSTM model which uses the word embeddings pretrained on Twitter data\footnote{\url{https://nlp.stanford.edu/projects/glove/}}.

\vspace{0.05in}
\noindent\textbf{Dense-LSTM} \cite{wu2018thu_ngn}: A state-of-the-art single-task sarcasm detection model ranked top-1 on the SemEval-18 dataset. The model is a densely connected LSTM network consisting of four Bi-LSTM layers and the word embeddings pretrained on two Twitter datasets. 

\vspace{0.05in}
\noindent\textbf{BERT}: We finetune the BERT model  \cite{devlin-etal-2019-bert} with an additional simple classifier directly on the target dataset.  

\vspace{0.05in}
\noindent\textbf{S-BERT} is a two-stage finetuning of the BERT model. We first finetune BERT on the source dataset and the best model is selected for further fine-tuning on the target dataset.

\vspace{0.05in}
\noindent\textbf{MTL}: We implemented a multi-task learning (MTL) model, which has the same architecture as LOANT except that the domain discriminator is removed. We use BERT as the shared text encoding network.

\vspace{0.05in}
\noindent\textbf{MTL+LO}: In this baseline, we applied latent optimization to MTL. As MTL does not have the adversarial discriminator, we use the domain-specific losses to optimize latent representations:
\begin{equation}
z_s^\prime = z_s - \gamma\pdv{\mathcal{L}_s(z_s)}{z_s}
\end{equation}
\begin{equation}
z_t^\prime = z_t - \gamma\pdv{\mathcal{L}_t(z_t)}{z_t}
\end{equation}
We use the above to replace Equations \ref{eq:loant-look-ahead-1} and \ref {eq:loant-look-ahead-2} and keep the rest training steps unchanged. This model is compared against MTL to study the effects of LO in non-adversarial training for cross-domain sarcasm detection.

\vspace{0.05in}
\noindent\textbf{ANT}: This is the conventional adversarial neural transfer model with the same architecture as LOANT. The only difference is that we do not apply latent optimization. For fair comparisons, we use BERT as the text encoder. 

\vspace{0.05in}
\noindent\textbf{ANT+MAML}:
In Section \ref{sec:rationale-for-loant}, we discussed the similarity between LO and MAML. Therefore, we create a baseline that uses a MAML-like strategy for encouraging the collaboration of different loss terms. Instead of optimizing the latent representation $z_s$ and $z_t$, we first take a SGD step in the parameter space of $w_b$,
\begin{equation}
w_b^\prime = w_b + \gamma\pdv{\mathcal{L}_d(z_s, z_t)}{w_b}.
\end{equation}
After that, we use $w_b^\prime$ to compute the gradients used in the actual updates to all model parameters, including $w_b$. 



\begin{table*}[t]
    \centering
     \small
    \begin{tabular}{lllll}
     & \multicolumn{4}{c}{Target: SemEval-18}\\
     \hline
     & Model & F-score & Recall & Precision \\
    \hline
        \multirow{5}{*}{Single-task}& Random$^{\dag}$ & 0.3730 & 0.3730 & 0.3730\\ 
        & Unigram SVM$^{\dag}$  & 0.5890 & 0.6590 & 0.5320 \\
        & LSTM$^{\dag}$ & 0.5260 & 0.4440 & 0.6450 \\
        & DenseLSTM $^{*}$ & \underline{0.6510} & 0.7106 & 0.6005 \\
        & BERT & 0.6626 & 0.7055 & 0.6246 \\ 
        \hline
        
       \multirow{6}{*}{Source: Pt{\'a}{\v{c}}e}
       & S-BERT & 0.6676 & 0.7055 & 0.6337  \\ 
       & MTL & 0.6404 & 0.7896 & 0.5386 \\
       & ANT & 0.6348 & 0.8187 & 0.5184 \\
       & MTL+LO & 0.6598 & 0.7346 & 0.5989 \\
       & ANT+MAML & 0.6454 & 0.7540 & 0.5641 \\
       & LOANT (ours) & \textbf{0.6702} & 0.8025 & 0.5754 \\
       \hline
       \multirow{6}{*}{Source: Ghosh}
       & S-BERT & 0.6512 & 0.7766 & 0.5607  \\ 
       & MTL & 0.6525 & 0.7475 & 0.5789 \\
       & ANT & 0.6626 & 0.8899 & 0.5278 \\
       & MTL+LO & 0.6622 & 0.8058 & 0.5620\\
       & ANT+MAML & 0.6338 & 0.7281 & 0.5610 \\
       & LOANT (ours) & \textbf{0.6818} & 0.7734 & 0.6096 \\
       \hline
    \end{tabular}
    \begin{tabular}{llll}
    \multicolumn{4}{c}{Target: iSarcasm} \\
     \hline
     Model & F-score & Recall & Precision \\
    \hline
        SIARN$^{\ddag}$ & 0.3420 & 0.7820 & 0.2190 \\ 
        MIARN$^{\ddag}$ & \underline{0.3640} & 0.7930 & 0.2360 \\
        LSTM$^{\ddag}$ & 0.3360 & 0.7470 & 0.2170 \\
        DenseLSTM$^{\ddag}$ & 0.3180 & 0.2760 & 0.3750 \\
        BERT & 0.3492 & 0.4904 & 0.2711 \\ 
         \hline
         S-BERT & 0.3710 & 0.5541 & 0.2788 \\ 
         MTL & 0.3767 & 0.3503 & 0.4074 \\
         ANT & 0.3857 & 0.5159 & 0.3079 \\
         MTL+LO & 0.4379 & 0.4267 & 0.4496 \\
         ANT+MAML & 0.3951 & 0.5605 & 0.2923 \\
         LOANT (ours) & \textbf{0.4642} & 0.4968 & 0.4357\\
         \hline
          S-BERT & 0.3383 & 0.5732 & 0.2400 \\
          MTL & 0.3838 & 0.5159 & 0.3056 \\
          ANT & 0.4063 & 0.4904 & 0.3468 \\
          MTL+LO & 0.3987 & 0.4012 & 0.3962 \\
          ANT+MAML & 0.3589 & 0.4904 & 0.2830 \\
          LOANT (ours) & \textbf{0.4101} & 0.4649 & 0.3668 \\
         \hline
    \end{tabular}
    \begin{tablenotes}
      \small
      \item $^{\dag}$ Results reported in \cite{van2018semeval}, $^{*}$ in \cite{wu2018thu_ngn} and $^{\ddag}$ in \cite{oprea-magdy:2020:isarcasm}.
    \end{tablenotes}.
    \caption{Performance on the sarcastic class reported by single-task and multi-task models on the same test sets. The best performed F-score on the four groups of transfer learning are in bold. The best single task learning results are underlined.}
    \label{tab:supervised}
\end{table*}

        
        


 

\subsection{Experimental Settings}

\noindent\textbf{Model Settings.}
For all models using the BERT text encoder, we use the uncased version of the BERT-base model and take the 768-dimensional output from the last layer corresponding to the \texttt{[CLS]} token to represent a sentence. The BERT parameters are always shared between domains. For other network components, we randomly initialize the dense layers and classifiers. To minimize the effect of different random initializations, we generate the same set of initial parameters for each network component and use them across all baselines wherever possible. 

The source dense layer, the shared dense layer, and the target dense layer are single linear layers with input size of 768 and output size of 768 followed by the tanh activation. The classifier in all models consists of two linear layers. The first linear layer has input size of 768$\times$2 (taking both shared and domain-specific features) and output size of 768 followed by the ReLU activation. The second linear layer has input size 768 and output size 2 for binary classification. After that we apply the softmax operation. More details can be found in Appendix \ref{appendix:Hyperparameters_and_Model_Initialization}.

\vspace{0.05in}
\noindent\textbf{Training Setting.} We optimize all models using Adam \cite{kingma2014adam} with batch size of 128. We tune the learning rate (LR) on the development set from 1e-5 to 1e-4 in increments of 2e-5. To objectively assess the effects of latent optimization (LO), we first find the best LR for the base models such as ANT and MTL. After that, with the best LR unchanged, we apply LO to ANT and MTL. We use the cosine learning rate schedule for all models. All models are trained for 5 epochs on Nvidia V100 GPUs with 32GB of memory in mixed precision. Due to the large model size and pretrained weights of BERT, 5 epochs are sufficient for convergence. 

\vspace{0.05in}
\noindent\textbf{Evaluation Metrics.}
Following \cite{wu2018thu_ngn,van2018semeval,oprea-magdy:2020:isarcasm}, we select and compare models using the F-score on the sarcastic class in each dataset. We additionally report the corresponding Recall and Precision. In all our experiments, we use the development set for model selection and report their performance on the test set. To evaluate the efficiency of LOANT versus MAML-based training, we also compare their required GPU memory and average training time in each epoch. We compare models on the target domain datasets. Additional multi-domain performance can be found in Appendix \ref{appendix:Multi_Domain_Performance}.

\subsection{Comparison with the States of the Art}
We compare LOANT with state-of-the-art methods on the SemEval-18 dataset \cite{van2018semeval} and the iSarcasm datast \cite{oprea-magdy:2020:isarcasm}. Table \ref{tab:supervised} presents the test performance of LOANT and all baseline models. Our LOANT model consistently outperforms all single-task baselines by large margins. In particular, LOANT outperforms MIARN by 10.02\% on iSarcasm \cite{oprea-magdy:2020:isarcasm} whereas the fine-tuned BERT achieved 1.48\% lower than MIARN. 
On SemEval-18, the fine-tuned BERT achieves better test performance than other four single-task baselines. The results indicate that fine-tuning BERT, a popular baseline, does not always outperform the traditional LSTM networks specifically designed for the task. We hypothesize that the large BERT model can easily overfit the small datasets used, which highlights the challenge of sarcasm detection.


\subsection{Transfer Learning Performance}
The middle and bottom sections of Table \ref{tab:supervised} present the test performance of six transfer learning models (S-BERT, MTL, ANT, MTL+LO, ANT+MAML, and LOANT) under four groups of transfer learning experiments. These models generally outperform the single-task models, demonstrating the importance of transfer learning. Among these, we have the following observations.

\begin{table}[t]
    \centering
     \small
    \begin{tabular}{llll}
     & & SemEval-18 & iSarcasm\\
     \hline
     & Model & RAM/Time & RAM/Time \\
    \hline
        
        \multirow{3}{*}{\shortstack[l]{Source:\\ Pt{\'a}{\v{c}}e}}
        & LOANT & 1.01x/2.41x & 1.01x/2.55x \\ 
        & MTL+LO & 1.01x/1.92x & 1.01x/1.91x \\
        & ANT & 1.00x/1.00x & 1.00x/1.00x\\
        & ANT + MAML & 1.99x/8.31x & 1.93x/10.2x \\
        \hline
        
        \multirow{3}{*}{\shortstack[l]{Source:\\ Ghosh}}
        & LOANT & 1.01x/2.44x & 1.01x/1.94x\\ 
        & MTL+LO & 1.01x/1.94x & 1.01x/1.89x \\
        & ANT & 1.00x/1.00x & 1.00x/1.00x \\
        & ANT + MAML & 1.99x/8.41x & 1.93x/10.7x\\

      \hline
    \end{tabular}
    \caption{Running time and maximum memory footprint for different transfer learning methods.}
    \label{tab:time-memory}
\end{table}

\vspace{0.05in}
\noindent\textbf{Effects of the Domain Discriminator.} The performance differences between MTL and ANT can be explained by the addition of the domain discriminator, which encourages the shared features under the source domain and the target domain to have the same distributions. In the four pairs of experiments, ANT marginally outperforms MTL by an average of 0.9\% F-score. In the \ptacek{} $\to$ SemEval-18 experiment, the domain discriminator causes F-score to decrease by 0.56\%. Overall, the benefits of the adversarial discriminator to transfer learning appear to be limited. As discussed earlier, the competition between the domain-specific losses and the negative domain discrimination loss may have contributed to the ineffectiveness of ANT. 

\vspace{0.05in}
\noindent\textbf{Effects of Latent Optimization.} We can observe the effects of LO by comparing ANT with LOANT and comparing MTL with MTL+LO. Note that in these experiments we adopted the best learning rates for the baseline models ANT and MTL rather than the latent-optimized models. On average, LOANT outperforms ANT by 3.42\% in F-score and MTL+LO outperforms MTL by 2.63\%, which clearly demonstrates the benefits provided by latent optimization. 

\vspace{0.05in}
\noindent\textbf{Latent Space vs. Model Parameter Space.} In the ANT+MAML baseline, we adopt a MAML-like optimization strategy, which performs the look-ahead in the BERT parameter space instead of the latent representation space. 
Interestingly, this strategy does not provide much improvements and on average performs 1.40\% worse than ANT. LOANT clearly outperforms ANT+MAML. 

In addition, optimization in the latent space also provides savings in computational time and space requirements. Table \ref{tab:time-memory} shows the time and memory consumption for different transfer learning methods. Adding LO to ANT has minimal effects on the memory usage, but adding MAML nearly doubles the memory consumption. On average, ANT+MAML increases the running time of LOANT by 3.1 fold.

\vspace{0.05in}
\noindent\textbf{The Influence of Domain Divergence.} In transfer learning, the test performance depends on the similarity between the domains. We thus investigate the dissimilarity between datasets using the Kullback–Leibler (KL) divergence between the unigram probability distributions,
\begin{equation}
    d_{KL} = \sum_{g\in V} P_t (g) \mathrm{log} \frac{P_t(g)}{P_s(g)}.
\end{equation}
where $P_s(g)$ and $P_t(g)$ are the probabilities of unigram $g$ for the source domain and target domain respectively. $V$ is the vocabulary. Table \ref{tab:KL} shows the results. \ptacek{} is more similar to the two target datasets than Ghosh. Among the two target datasets, iSarcasm is more similar to \ptacek{} than SemEval-18. 

Comparing LOANT and ANT, we observe that the largest improvement, 7.85\%, happens in the \ptacek{} $\to$ iSarcasm transfer where domain divergence is the smallest. The \ptacek{} $\to$ SemEval-18 transfer comes in second with 3.54\%. Transferring from Ghosh yields smaller improvements. Further, we observe the same trend in the comparison between MTL+LO and MTL. The largest improvement brought by LO is 6.12\% in the \ptacek{} $\to$ iSarcasm transfer. As one may expect, applying LO leads to greater performance gains when the two domains are more similar.


\begin{table}[t]
    \centering
    \begin{tabular}{c c c}
         & SemEval-18 & iSarcasm \\
         \hline
         Pt{\'a}{\v{c}}ek & 0.1631 & 0.0521 \\
         Ghosh & 0.2300 & 0.2217 \\
         \hline
    \end{tabular}
    \caption{The KL divergence of word probability over the overlapped vocabulary for each pair of domains.}
    \label{tab:KL}
\end{table}

\section{Conclusion}

Transfer learning holds the promise for the effective utilization of multiple datasets for sarcasm detection. In this paper, we propose a latent  optimization (LO) strategy for adversarial transfer learning for sarcasm detection. By providing look-ahead in the gradient updates, the LO technique allows multiple losses to accommodate each other. This proves to be particularly effective in adversarial transfer learning where the domain-specific losses and the adversarial loss potentially conflict with one another. With the proposed LOANT method, we set a new state of the art for the iSarcasm dataset. We hope the joint utilization of multiple datasets will contribute to the creation of contextualized semantic understanding that is necessary for successful sarcasm detection. 


\section*{Acknowledgments}
This research is supported by the National Research Foundation, Singapore under its the AI Singapore Programme (AISG2-RP-2020-019), NRF Investigatorship (NRF-NRFI05-2019-0002), and NRF Fellowship (NRF-NRFF13-2021-0006); the Joint NTU-WeBank Research Centre on Fintech (NWJ-2020-008); the Nanyang Assistant/Associate Professorships (NAP); the RIE 2020 Advanced Manufacturing and Engineering Programmatic Fund (A20G8b0102), Singapore; NTU-SDU-CFAIR (NSC-2019-011). Any opinions, findings and conclusions or recommendations expressed in this material are those of the authors and do not reflect the views of the funding agencies.

\bibliography{custom}
\bibliographystyle{acl_natbib}

\newpage

\appendix
\begin{center}\large\bfseries
Appendix for ``Latent-Optimized Adversarial Neural Transfer for Sarcasm Detection''
\end{center}

\section{First-order Approximation}
\label{appendix:First_order_Approximation}

Here we explain the gradients for the model parameters $w_b, w_{sh}, \phi_{s}, \phi_{t}$ and $\theta_d$. 
Generically, we apply the first-order approximation by substituting Eq.\ref{expansion-2} into Eq. \ref{expansion-1} and setting the Hessian to zero, which gives
\begin{equation}
\begin{split}
\odv{\mathcal{L}^{\text{LO}}}{w} = \pdv{\mathcal{L}^{\text{LO}}}{w} +  \pdv{\mathcal{L}_{s}(z_{s}^\prime)}{z_{s}^\prime}\pdv{z_{s}}{w} + \pdv{\mathcal{L}_{t}(z_{t}^\prime)}{z_{t}^\prime}\pdv{z_{t}}{w}.
\end{split}
\end{equation}
Note that $z_s$ and $z_t$ depend on only the parameter $w_b$. For the rest of the parameters, $w_{sh}, \phi_s, \phi_t$ and $\theta_d$, the partial derivatives $\pdv{z_s}{w}$ and $\pdv{z_t}{w}$ are zero. 

Now we consider the joint objective (Eq. \ref{eq:new_joint_loss}). The total derivative of $\mathcal{L}^{\text{LO}}$ against $w_b$ is
\begin{equation}
\begin{split}
\odv{\mathcal{L}^{\text{LO}}}{w_b} = \; & \pdv{\mathcal{L}_{s}(z_s^{\prime})}{w_b} + \pdv{\mathcal{L}_{t}(z_t^{\prime})}{w_b} + \pdv{\mathcal{L}_{s}(z_{s}^\prime)}{z_{s}^\prime}\pdv{z_{s}}{w_b} \\
& + \pdv{\mathcal{L}_{t}(z_{t}^\prime)}{z_{t}^\prime}\pdv{z_{t}}{w_b} - \pdv{\mathcal{L}_d(z_s, z_t)}{w_b} \\
\end{split}
\end{equation}

For the rest of the parameters, the computation is slightly different as they do not contribute to $z_s$ and $z_t$. Thus, $\pdv{z_s}{w}=0$ and $\pdv{z_t}{w}=0$.
\begin{equation}
\begin{split}
\odv{\mathcal{L}^{\text{LO}}}{w_{sh}}=
\pdv{\mathcal{L}^{\text{LO}}}{w_{sh}}
= \; & \pdv{\mathcal{L}_{s}(z_s^{\prime})}{w_{sh}} + \pdv{\mathcal{L}_{t}(z_t^{\prime})}{w_{sh}} \\ & - \pdv{\mathcal{L}_d(z_s, z_t)}{w_{sh}} 
\end{split}
\end{equation}
Besides, $\phi_s$ is only updated by $\mathcal{L}_s(z_s^\prime)$ while $\phi_t$ is only updated by $\mathcal{L}_t(z_t^\prime)$. Thus, we have
\begin{align}
\begin{split}
\odv{\mathcal{L}^{\text{LO}}}{\phi_s}= \pdv{\mathcal{L}_s(z_s^\prime)}{\phi_s}
\end{split}\\
\begin{split}
\odv{\mathcal{L}^{\text{LO}}}{\phi_t}=
\pdv{\mathcal{L}_t(z_t^{\prime})}{\phi_t}
\end{split}
\end{align}
The parameter of the domain discriminator $\theta_d$ is updated to minimize $\mathcal{L}_d(z_s, z_t)$. This is in contrast to the rest of the model, which minimizes $- \mathcal{L}_d(z_s, z_t)$. The update rule for $\theta_d$ is
\begin{equation}
\begin{split}
\theta_d \leftarrow \theta_d - \eta \pdv{\mathcal{L}_d(z_s, z_t)}{\theta_d}
\end{split}
\end{equation}

\section{Hyperparameters and Model Initialization}
\label{appendix:Hyperparameters_and_Model_Initialization}


We set the batch size to 128 for all models and search for the optimal learning rate (LR) from 2e-5 to 1e-4 in increments of 2e-5 using the F-score on the development set. We show the best learning rates found in Table \ref{grid-search-results}. 

The best learning rate for fine-tuning BERT on SemEval-18 and iSarcasm is 4e-5. S-BERT model is finetuned twice, first on the source domain and then on the target domain. Thus, we search for one best learning rate for each finetuning using the source and target development sets respectively. The best first-round LR is 6e-05 for Pt{\'a}{\v{c}}e and 8e-5 for Ghosh. 

Other models, MTL, ANT and the LO-adpated versions are selected using the target development set. For a rigorous comparison, we use the best LR for ANT when training LOANT and the best LR for MTL when training MTL+LO.

We follow the released code\footnote{\url{https://github.com/fungtion/DANN}} to implement the Gradient Reversal Layer. It is controlled by a schedule which gradually increases the weight of the gradients from the domain discrimination loss. 

\section{Source Domain Performance}
\label{appendix:Multi_Domain_Performance}

The original goal of the paper is to use automatically collected sarcasm datasets, which are large but \textit{noisy}, to improve performance on human-annotated datasets, which are \textit{clean} and provide good performance measure. That is why we provided only the target domain performance. 

Upon close inspection, LOANT also improves the performance on the source domain, even though model selection was performed on the target domain. Table \ref{both-results-selectedbyTarget}shows the results.

In Table \ref{both-results-selectedbyBoth}, we also show the results after model selection on both domains. Naturally, this might lead to slightly lowered target-domain performance than achieved by model selection on target domain only. Comparing LOANT with ANT, and MTL+LO with MTL, our results show that, in most cases, LO-based models improve both source and target domain F1. In particular, target domain F1 obtains more improvement than source domain F1. This suggests that LO provides benefits to knowledge transfer. 

\begin{table}[t]
\small
    \centering
    \begin{tabular}{c|c|c|c|c}
    \hline
    Models & \shortstack[l]{Pt{\'a}{\v{c}}e $\to$\\ SemEval} & \shortstack[l]{Ghosh $\to$\\ SemEval} & \shortstack[l]{Pt{\'a}{\v{c}}e $\to$\\ iSarcasm} & \shortstack[l]{Ghosh $\to$\\ iSarcasm} \\
    \hline
    S-BERT & 1e-4 & 1e-4 & 4e-5 & 2e-5 \\
    MTL & 6e-4 & 8e-5 & 4e-5 & 1e-4 \\
    MTL+LO & 6e-4 & 8e-5 & 4e-5 & 1e-4\\
    ANT & 2e-5 & 4e-5 & 2e-5 & 2e-5 \\
    ANT+MAML & 2e-5 & 4e-5 & 2e-5 & 2e-5 \\
    LOANT & 2e-5 & 4e-5 & 2e-5 & 2e-5 \\
    \hline
    \end{tabular}
    \caption{Learning rate chosen by each model on the given search grid.}
    \label{grid-search-results}
\end{table}

\begin{table}[t]
\small
    \centering
    \begin{tabular}{c|c|c|c|c}
    \hline
    Domain & ANT & LOANT & MTL & MTL+LO \\
    \hline
    \hline
    Ptacek & 0.8307 & \textbf{0.8484} & \textbf{0.8640} & 0.8629 \\
    iSarcasm & 0.3857 & \textbf{0.4642} & 0.3767 & \textbf{0.4379} \\
    \hline
    Average &  0.6082 & \textbf{0.6563} & 0.62035 & \textbf{0.6504} \\
    \hline
    \hline
    Ghosh & \textbf{0.7345} & 0.6596 & 0.6609 & \textbf{0.6688} \\
    iSarcasm & 0.4063 & \textbf{0.4101} & 0.3838 & \textbf{0.3953} \\
    \hline
    Average & \textbf{0.5704} & 0.5349 & 0.5224 & \textbf{0.5321} \\
    \hline
    \hline
    Ptacek & \textbf{0.8626} & 0.8612 & \textbf{0.8722} & 0.8666 \\
    SemEval18 & 0.6348 & \textbf{0.6702} & 0.6404 & \textbf{0.6598} \\
    \hline
    Average & 0.7487 & \textbf{0.7657} & 0.7563 & \textbf{0.7632} \\
    \hline
    \hline
    Ghosh & 0.7161 & \textbf{0.7752} & \textbf{0.7700} & 0.7579 \\
    SemEval18 & 0.6626 & \textbf{0.6818} & 0.6525 & \textbf{0.6622} \\
    \hline
    Average & 0.6894 & \textbf{0.7285} & \textbf{0.7113} & 0.7101 \\
    \hline
    
    \end{tabular}
    \caption{Test F1 score for both domains using model selection on the target domain only. }
    \label{both-results-selectedbyTarget}
\end{table}

\begin{table}[!t]
\small
    \centering
    \begin{tabular}{c|c|c|c|c}
    \hline
    Domain & ANT & LOANT & MTL & MTL+LO \\
    \hline
    \hline
    Ptacek 	& 0.8307	  & \textbf{0.8484}	 & \textbf{0.8640}	& 0.8629\\
    iSarcasm  & 0.3857	  & \textbf{0.4642}	 & 0.3767	& \textbf{0.4379} \\
    \hline
    Average &  0.6082 & \textbf{0.6563} & 0.6204 & \textbf{0.6504}\\
    \hline
    \hline
    Ghosh 	& 0.7787	  & \textbf{0.7826}	 & \textbf{0.7859}	& 0.7807 \\
    iSarcasm  & \textbf{0.3965}	  & 0.3215	 & 0.3764	& \textbf{0.3953} \\
    \hline
    Average & \textbf{0.5876} & 0.5521	& \textbf{0.5812}	& 0.5880\\
    \hline
    \hline
    Ptacek 	& 0.8567	  & \textbf{0.8612}	 & \textbf{0.8720}	& 0.8632 \\
    SemEval18 & 0.6463	  & \textbf{0.6702}	 & 0.6594	& \textbf{0.6666} \\
    \hline
    Average & 0.7515 & \textbf{0.7657} & \textbf{0.7657} & 0.7649 \\
    \hline
    \hline
    Ghosh 	& 0.7919	  & \textbf{0.7962}	 & 0.7672	& \textbf{0.7884} \\
    SemEval18 & 0.6427	  & \textbf{0.6490}	 & 0.6357	& \textbf{0.6442} \\
    \hline
    Average & 0.7173 & \textbf{0.7226} & 0.7015 & \textbf{0.7163}\\
    \hline
    
    \end{tabular}
    \caption{Test F1 score for both domains using model selection on the average F1 of the two domains.}
    \label{both-results-selectedbyBoth}
\end{table}


\end{document}